\documentclass[11pt]{article}

\usepackage[preprint]{acl}

\usepackage{times}
\usepackage{latexsym}

\usepackage[T1]{fontenc}

\usepackage[utf8]{inputenc}

\usepackage{microtype}

\usepackage{inconsolata}

\usepackage{graphicx}

\usepackage{float}
\title{Lost in Historical Time? \\ A Polish History \textit{Matura} Benchmark for Large Language Models}

\author{
  Adrian Trzoss\textsuperscript{1,6}\thanks{\;Correspondence: \texttt{adrian.trzoss@amu.edu.pl}} \quad
  Kacper Dudzic\textsuperscript{1,2,3} \quad
  Wiktor Werner\textsuperscript{1} \quad
  Marcin Moskalewicz\textsuperscript{1,4,5} \\
  \\
  \textsuperscript{1}Adam Mickiewicz University, Poznań, Poland \\
  \textsuperscript{2}IDEAS Research Institute, Warsaw, Poland \\
  \textsuperscript{3}AMU Center for Artificial Intelligence, Poznań, Poland \\
  \textsuperscript{4}Poznań University of Medical Sciences, Poznań, Poland \\
  \textsuperscript{5}Maria Curie-Skłodowska University, Lublin, Poland \\
  \textsuperscript{6}WSB Merito University, Poznań, Poland \\
  \rule{0pt}{3ex}
}

\begin{document}
\maketitle
\begin{abstract}
Language models are widely used by students as knowledge sources, yet benchmarks rarely assess their interpretative historical reasoning. 
We evaluate eight leading LLMs on the Polish high school exit exam (Matura) in history---three official papers from 2023--2025, comprising short-answer questions and extended essays--- and compare model performance against the human examinee population. 
Although models score near the ceiling, aggregate scores mask distinct competency profiles: rankings are unstable across task types, source modalities, and geographical scopes, with a consistent penalty for Polish versus Global history content.
Qualitative error analysis reveals two recurring failure modes---\textit{source decontextualization}, when models reason from source content rather than treating it as an object of analysis, and \textit{temporal disorientation}, when responses are historically misplaced. 
This study introduces the first LLM history benchmark grounded in Polish national curriculum \footnote{We release our code at: \url{https://anonymous.4open.science/r/history-matura-llm-evaluation-39B4/.}}.
\end{abstract}

\section{Introduction}

AI chatbots powered by Large Language Models have become a significant source of knowledge in educational settings, widely adopted by secondary school and university students alike. 
A growing body of research \cite{gilson2023does,locatelli2024examining,hendrycks2020measuring,dargis2024evaluating} evaluates these models with respect to knowledge and pedagogical value using standardized benchmarks.
However, more interpretative assessments of historical topics, which rarely admit clear-cut answers, remain substantially underrepresented. 
This paper evaluates the performance of frontier LLMs on the Polish high school exit exam (\textit{Matura}) in history, administered by the state-run Central Examination Board (\textit{Centralna Komisja Egzaminacyjna}, CKE) \footnote{\url{https://cke.gov.pl/egzamin-maturalny/egzamin-maturalny-w-formule-2023/}}. 
We test eight models on official examination papers from 2023--2025 and assess performance along two dimensions: comparing models against one another, and against human examinees.
Matriculation examinations in history are a methodologically valid benchmark of historical skills because they assess both factual knowledge and historical reasoning within the national curriculum framework.
Moreover, the question pool changes annually, and tasks are not designed with LLMs in mind, reducing research design bias. 
Finally, each paper is accompanied by a standardized marking scheme with model answers and grading criteria, enabling objective scoring. 
As a result, the benchmark represents a rare instance of an open-ended humanities task with an externally defined and relatively objective evaluation framework.

We investigate whether LLMs outperform human examinees overall, whether performance varies systematically by question type, source modality, and historical scope---including a predicted penalty on Polish versus Global history content \cite{dadas2025evaluating}. 
Additionally, we examine whether certain task formulations prove systematically incomprehensible under baseline prompting conditions.
We situate this study in relation to two bodies of prior work: global history benchmarks, which have not addressed non-Anglophone educational content, and Polish-language NLP benchmarks, which have not addressed open-ended humanities tasks.

\section{Related Work}

Hauser introduced HiST-LLM benchmark \cite{hauser2024large}, demonstrating substantial variation in model performance across historical regions and periods. 
Chartier extended this line of work with HiBenchLLM \cite{chartier2025hibenchllm}, confirming that models perform worse on non-Anglophone content---including French. 
These benchmarks, however, rely on arbitrarily constructed question sets and manually designed evaluation criteria, and neither relates model performance to human baselines.
Within the Polish context, LLMzSzŁ (LLMs Behind the School Desk) by \citet{jassem2025llmzsz} is the first large-scale LLM benchmark for high school exams, built from archival examinations across multiple subjects. The finding that multilingual models frequently outperform monolingual ones, though the latter can be competitive under size constraints, do not transfer directly to our study: LLMzSzŁ relies exclusively on closed multiple-choice questions, which are far easier to evaluate automatically and at scale, as well as without history coverage. We address both gaps simultaneously: ours is the first benchmark to evaluate LLMs on Polish historical content using authentic open-ended examination tasks, and the first to compare model performance against a human distributional baseline---the two limitations that characterize prior work in this space.

\section{Dataset}

The benchmark consists of three official history \textit{Matura} papers in the new 2023, 2024, and 2025 exam formats.
Each contains \emph{tasks}---thematic units built around source materials---each of which may include one or more \emph{questions}. 
The 2023 paper included 36 questions across 25 tasks; the 2024 paper, 39 questions across 25 tasks; and the 2025 paper, 37 questions across 24 tasks. 
The final task in each paper required an essay of at least 600 words on one of three proposed topics.
All papers covered a broad chronological range from antiquity to the late twentieth century, with roughly equal coverage of Polish and Global history.
Short-answer questions varied widely in format: single-word or single-phrase responses (naming a figure or event), multiple-choice items, binary judgment tasks (true/false) with brief justificatory reasoning, and source-analysis exercises (3--5 sentences). 
Each \emph{task}---except the essay---included source materials, which appeared in varying combinations: historical texts, iconographic materials (maps, photographs), and tables (economic or genealogical). 
Most questions were worth one or two points; the maximum total score per paper was 60 points, with 15 allocated to the essay.
Responses were scored according to official CKE grading guidelines and cross-checked by a high-level human expert---a CKE-trained and experienced \textit{Matura} examiner.

\section{Methodology}

\subsection{Model Selection}

We evaluated a total of $N = 8$ models from 4 leading providers: GPT-4o~\cite{openai2024gpt4ocard} and GPT-5.4~\cite{gpt54} from OpenAI, Claude Sonnet 3.7~\cite{sonnet37} and Claude Sonnet 4.6~\cite{sonnet46} from Anthropic, Gemini 2.5 Pro~\cite{comanici2025gemini25pushingfrontier} and Gemini 3.1 Pro~\cite{gemini31} from Google, as well as Grok 4~\cite{grok4} and Grok 4.20 from xAI. 

Our model selection criteria encompassed public interest and general performance, the ability to handle image inputs, availability on OpenRouter\footnote{\url{https://openrouter.ai/}}, and zero-barrier availability via free web interfaces (methodologically motivated by real-world educational deployment, where high school students interact with cloud-hosted web chat interfaces rather than locally deployed open-weight checkpoints).

\subsection{Model Inference Procedure}

Model outputs were obtained via the OpenRouter API. 
Each question was passed in a separate API call. 
The question text was passed along with the prompt (all in Polish), whereas the image inputs were included in an additional API call payload. 

Each short-answer question was passed to each model 3 times with the default temperature setting to evaluate potential inconsistencies arising from the non-deterministic nature of the models. 
Similarly, each essay topic was passed 3 times, with all possible topics from a sheet evaluated for each model (in contrast to a student who must choose one); this adjustment allowed for a more comprehensive evaluation of model performance on the essay section---topics concern various time periods and problems, and a model is not guaranteed to perform equally well on each.
In total, 2688 API calls for short answer questions (36 questions $\times$ 25 tasks from the 2023 sheet, 39 $\times$ 25 from 2024, and 37 $\times$ 24 from 2025; each question set $\times$ 3 attempts $\times$ 8 models each) and 216 for essays (3 essay topics $\times$ sheet $\times$ 3 attempts $\times$ 8 models each) have been made. 
We use a zero-shot inference without in-context examples to measure baseline out-of-the-box robustness, mirroring casual student usage.

\subsection{Data Preparation and Annotation}

Each question was manually annotated along three dimensions: geographical scope (Polish vs. Global history); historical period (antiquity, medieval, early modern, nineteenth century, twentieth century, Polish People's Republic); and source material type (text only, photo only, photo and text, table-based).

\subsection{Quantitative Analysis}

Three overall model rankings were computed---all tasks combined, short-answer only, and essays only---based on mean normalized scores aggregated across all years and runs, with uncertainty estimated via 95\% bootstrap confidence intervals (5,000 replications). 
Rankings were further aggregated by each annotation category. 
For human vs. model comparisons, we calculated the Wasserstein distance and equating (see Appendix~\ref{sec:appendix_d}). 
Because all three essay topics are evaluated for each model---compared with only one topic for human examinees---a complete model run yields a maximum of 90 points, whereas the human is 60.
All comparisons, therefore, use normalized scores to account for this asymmetry (see Section~\ref{sec:limitations}).

\section{Results}

\subsection{Short-answer questions vs. Essays}

\begin{figure}[t]
  \includegraphics[width=\columnwidth]{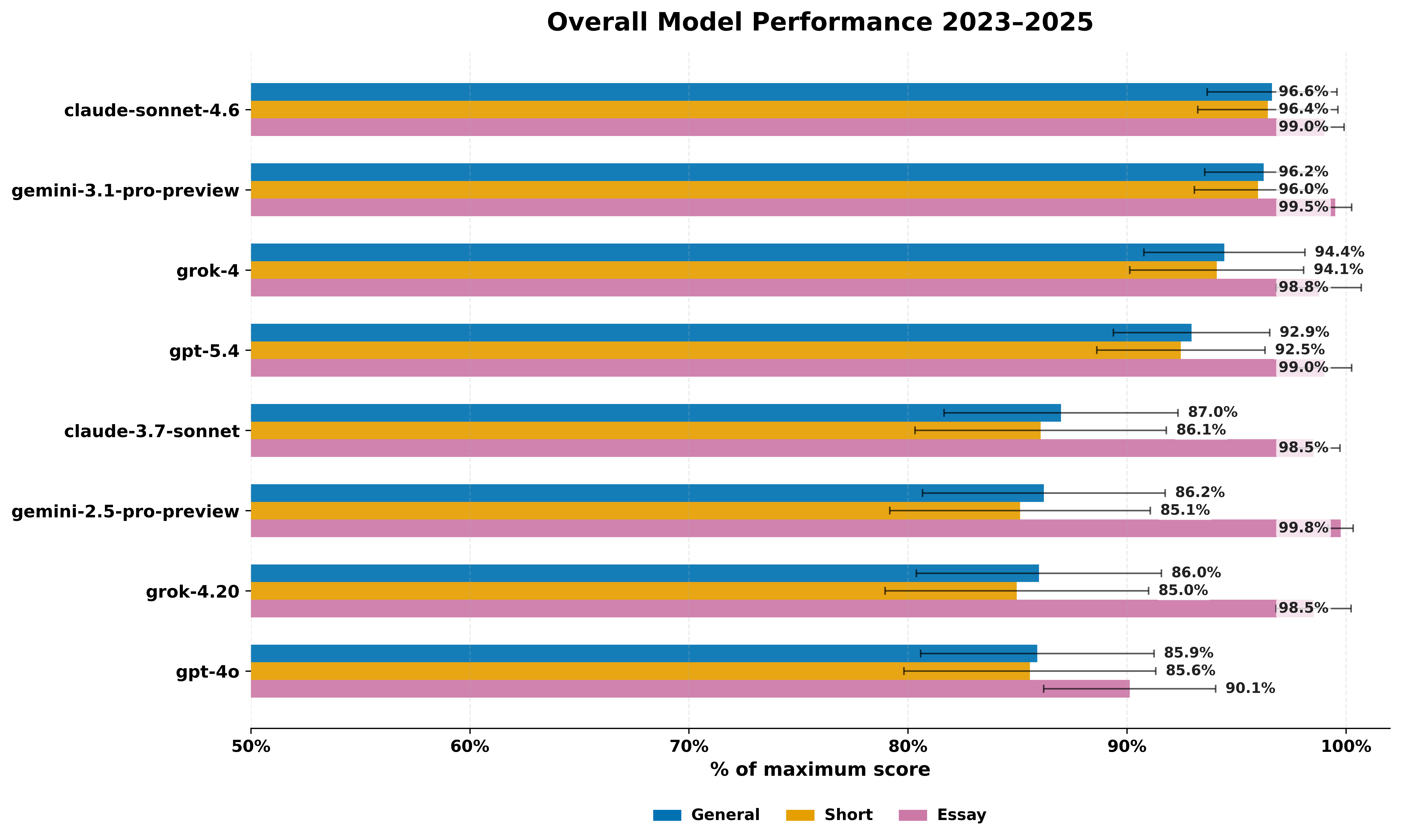}
  \caption{Overall performance of LLMs based on normalized scores aggregated across all years and runs: all tasks combined, short-answer questions, and essays only. Error bars represent 95\% bootstrap CI.
}
  \label{fig1}
\end{figure}

Figure~\ref{fig1} reveals a three-tier structure: Claude Sonnet 4.6 and Gemini 3.1 Pro lead (96.6\% \& 96.2\%), Grok 4 and GPT-5.4 form a semi-middle tier, and the remaining models cluster with indistinguishable confidence intervals. Short-answer performance is the primary discriminator between models.
Essay scores are uniformly high across seven models, providing no discriminative power on the benchmark; GPT-4o is the sole exception at 90.1\%.
The aggregate ranking, however, masks substantial instability across topical and source-type categories---and remains far above the human examinee average of 44.1\% across all analyzed years (Table~\ref{tab:human_model_bootstrap_full}).

\subsection{Polish vs. Global History}

\begin{figure}[t]
  \includegraphics[width=\columnwidth]{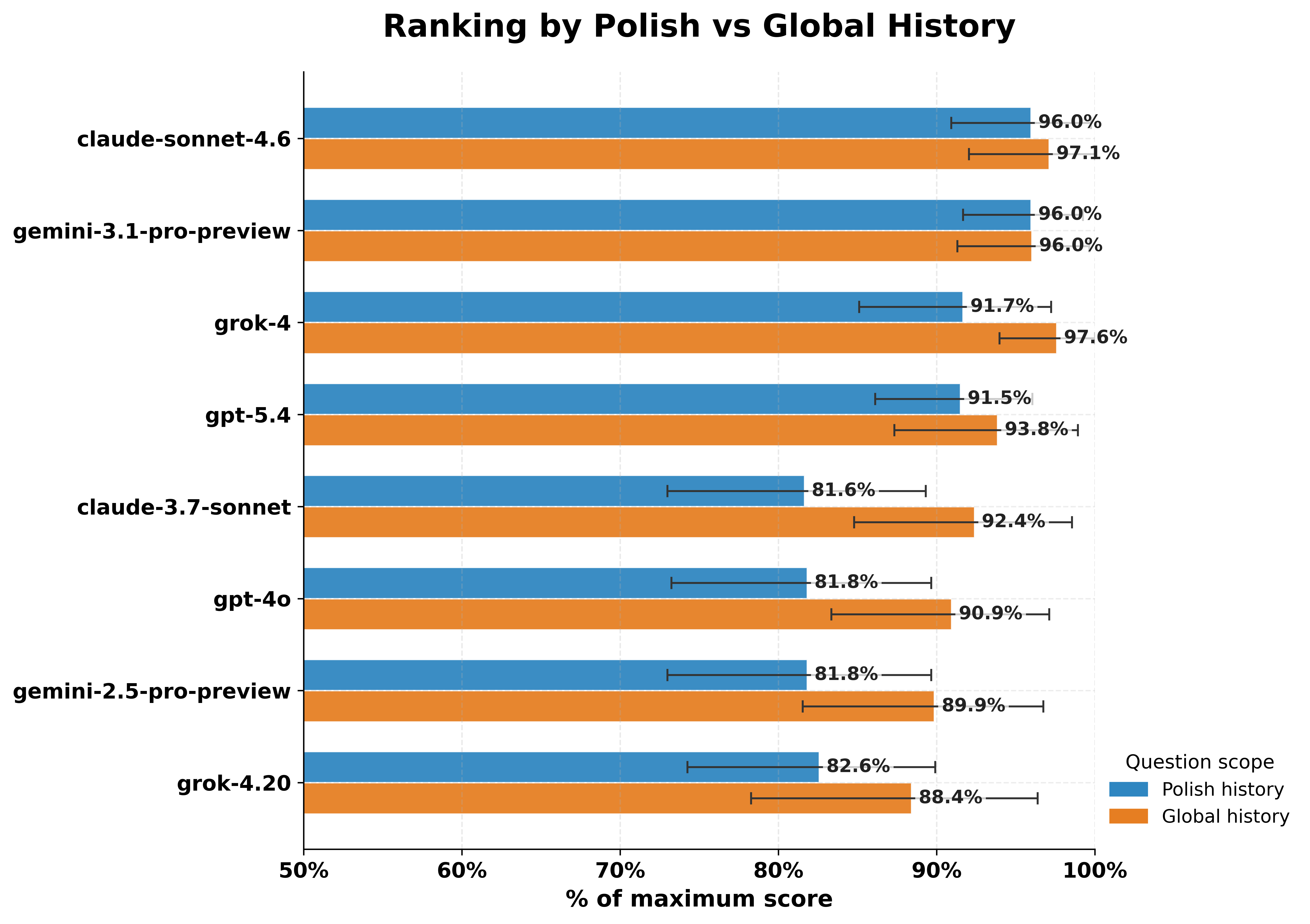}
  \caption{Model rankings by geographical scope (short-answer questions). Error bars represent 95\% bootstrap confidence intervals.
}
  \label{fig2}
\end{figure}

Figure~\ref{fig2} shows model rankings split by geographical scope.
Almost all models score higher on Global than on Polish history tasks, with a penalty on Polish content ranging up to 10.8 percentage points.
The gap is negligible for the top two models but substantial for the remaining six, where it ranges from 2.3 to 10.8 percentage points---suggesting that weaker models are disproportionately disadvantaged by nationally specific content.
The split produces non-trivial rank reordering: Grok 4, ranked third overall, scores 97.6\% on Global but drops to 91.7\% on Polish, falling behind GPT-5.4.
The epoch breakdown (Figure~\ref{fig4}) mirrors this pattern, with twentieth-century and Polish People's Republic (PRL) period showing the largest score variance.

\subsection{Source Type for Question Results}

\begin{figure}[t]
  \includegraphics[width=\columnwidth]{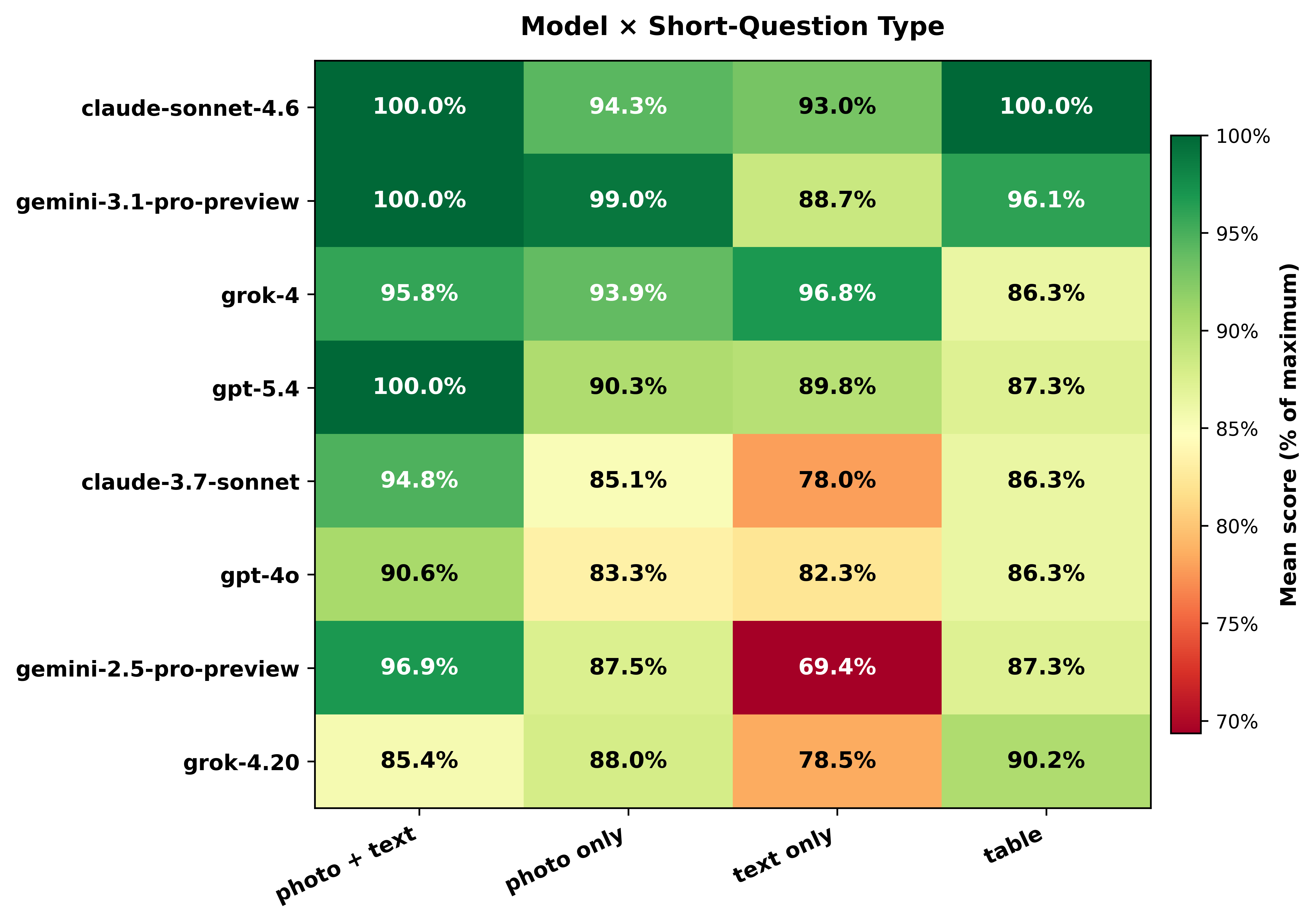}
  \caption{Mean normalized scores in the model $\times$ source type matrix.
  Photo + Text tasks yield the highest and most consistent scores; Text-only tasks show the widest variance.
  Error bars omitted for clarity.
}
  \label{fig3}
\end{figure}

Figure~\ref{fig3} presents mean normalized scores across the model $\times$ source-type matrix. 
Photo + Text tasks yield the highest and most consistent scores, suggesting a ceiling or redundant-cue effect.
Text-only tasks show the widest cross-model variance (69.4\%–96.8\%), with Gemini 2.5 Pro collapsing to the lowest cell in the matrix.
Gemini 3.1 Pro leads in Photo-only (99.0\%), Claude Sonnet 4.6 in Table-based tasks (100\%), and Grok 4 in Text-only (96.8\%), despite ranking third overall.
GPT-4o underperforms frontier models across all types.

\subsection{Failure Modes and Alignment Bias}

From 2688 model outputs, we manually examined the subset of questions for which models scored 0 points.
Table~\ref{tab2} lists the three hardest and most discriminating questions for models.
They are situated mainly in the 2024 and 2025 papers and concern either temporally oriented reasoning or culturally specific Polish content from the twentieth century and the PRL period.
Question \texttt{11\_02\_2025} is both in the hardest and most model-discriminating items in the benchmark:
"Determine which of the documents cited in fragments A–C was created first. Justify your answer by referring to the sources and your own knowledge" (see Appendix~\ref{sec:appendix_e}).
Models correctly recognized source names but failed to order them chronologically.
Both Gemini and Grok 4 scored maximum points across all runs; GPT-5.4 scored 1 point only in one run, while the others scored 0 across all three tests.
A cross-population comparison reveals an inversion: per CKE reports, human-hardest items disproportionately involve Photo + Text combinations (Table~\ref{tab:hardest_humans})---precisely the question type on which models perform most consistently (Figure~\ref{fig3}).


Manual inspection of all failed questions (Table~\ref{tab:zero-instances}) reveals two recurring failure modes. 
We call the first \textit{source decontextualization} since it occurs when a model reasons from the semantic content of a provided source instead of treating it as an object of historical analysis \cite{basmov2024llms}. 
E.g. in question \texttt{07\_2024}~(Table \ref{tab:benchmark_questions}), models consistently inferred chronological order from content comparison rather than authorial context, effectively treating sources as evidence about the world rather than as historically situated documents. 
This was observed across six models on all test runs; the only exceptions were Gemini 3.1 Pro and GPT-5.4.
The second mode, which we call \textit{temporal disorientation}, concerns questions requiring models to identify or order sequentially causally linked events, or locate them within a specific Polish historical period. 
E.g. on task \texttt{22\_2025}, models produced partially plausible responses---correctly identifying relevant actors or concepts---but placed them in the wrong period or order \cite{herel2024time,fatemi2025test}.
This pattern was particularly pronounced on the PRL-period questions (questions \texttt{23\_2024}, \texttt{25\_2024}).

The near-uniform essay ceiling masks a deeper problem of argumentative alignment.
A stance analysis of topics requiring counter-argumentation (Table~\ref{tab:model-stance}) reveals a systematic bias: models default to agreement with the thesis even when a counter-argument is expected.
GPT-5.4 failed to counter in 66.7\% of such runs; Grok-4.20 and Claude-3.7 showed the strongest resistance, correctly countering in 55.6\% and 77.8\% of cases respectively.
Yet, essays arguing the wrong stance scored just 0.01 points below those that argued correctly (14.68 vs. 14.69/15), confirming that the CKE grading scheme is nearly insensitive to the direction of argument. In other words, essay tasks reward textual fluency over genuine historical thinking and should be treated with caution as potential model alignment bias.

\section{Conclusions}

This first benchmark evaluation of LLMs on the Polish history \textit{Matura} shows that every model places at or above the 94th centile of the human cohort, yet aggregate scores mask substantial instability across task type, source modality, and geographical scope.
The Polish versus Global history split produces systematic rank reordering, extending Chartier's findings on non-Anglophone content to a Central-Eastern European context \cite{chartier2025hibenchllm}.
The identified failure modes suggest that what models lack on the hardest questions is not factual coverage but historically situated reasoning---the ability to reason within a period rather than merely about it. Nationally grounded, human-referenced benchmarks complement synthetic evaluation frameworks, particularly for languages and domains where global models remain undertested. 
Passing an examination is not the same as understanding its subject.

\section{Limitations}
\label{sec:limitations}

The benchmark comprises three examination papers, which limits statistical power and may not capture the full range of question types across years.
Official CKE reports provide only aggregate human score distributions rather than individual-level data, constraining the precision of human-model comparisons. 
A potential concern regarding human-AI comparability is that human candidates strategically select a single, most favorable essay topic, whereas models answered all three.
However, our empirical findings across multi-run evaluations show that frontier models (Gemini, Claude, GPT-5.4, and Grok families) achieve uniformly near-ceiling scores with minimal or zero variance across all prompts (see Appendix~\ref{sec:appendix_f}).

All evaluated models are closed-source, precluding analysis of the behavioral and architectural factors underlying observed failure modes. 
Potential contamination cannot be ruled out, as examination papers and official answers are publicly available online and may appear in pretraining corpora. 
Contamination risk is asymmetric across years: the 2023 paper has had the longest exposure window, yet per-year score variance is modest. 
Moreover, the benchmark's hardest questions---those requiring culturally specific PRL-period reasoning---are precisely where models score lowest, contrary to what direct memorization would predict.

All models were prompted under baseline conditions without chain-of-thought or few-shot examples, meaning results reflect but one point in a broader prompting space. 
While prompt engineering might alter individual model outputs, our objective is to assess baseline system defaults rather than upper-bound optimization; notably, models with native reasoning capabilities remained free to leverage internal reasoning tokens without prompt intervention.
Finally, no Polish monolingual model was included: Bielik, the most capable Polish-language model \cite{dadas2025evaluating}, lacks multimodal support and could not be evaluated on multimodal tasks.

\section{Ethical Considerations}

This work presents a methodological contribution and does not involve human subjects or personally identifiable information. 
All experiments were conducted using publicly available resources under their respective licenses.
The human expert---a CKE-trained and experienced \textit{Matura} examiner---worked voluntarily.

\bibliography{custom}

\onecolumn
\appendix
\counterwithin{figure}{section}
\counterwithin{table}{section}

\setcounter{figure}{0}
\renewcommand{\thefigure}{\thesection.\arabic{figure}}
\setcounter{table}{0}
\renewcommand{\thetable}{\thesection.\arabic{table}}

\section{Model Performance by Historical Epoch}
\label{sec:appendix_a}

\begin{figure}[H]
\centering
  \includegraphics[width=\columnwidth]{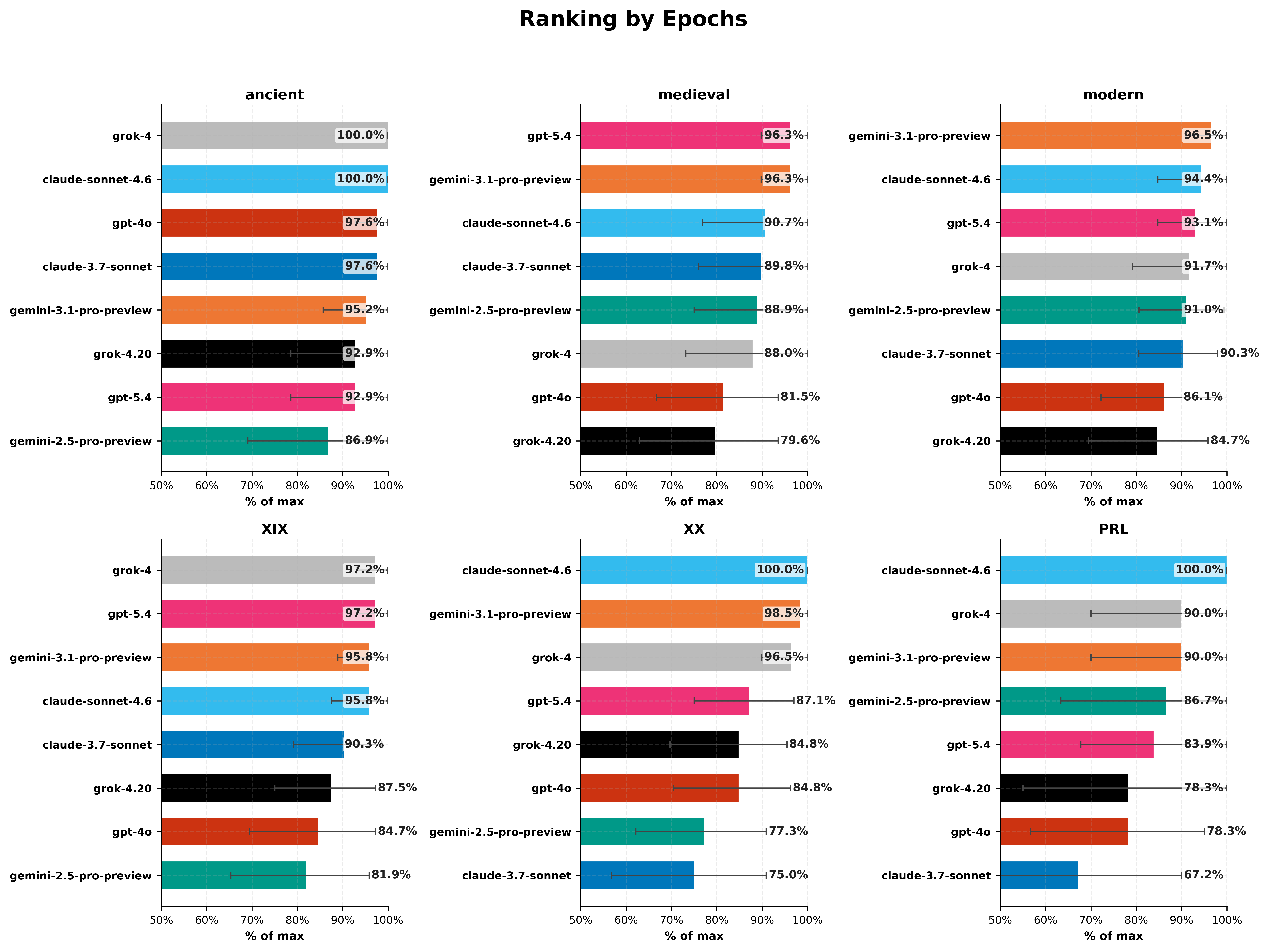}
  \caption{Model rankings by historical epoch (short-answer questions only). Aggregate benchmark rankings conceal substantial chronological specialization: performance is near-ceiling on antiquity tasks, but diverges markedly on twentieth-century and PRL-period content. The largest variance occurs on modern Polish history. Error bars represent 95\% bootstrap confidence intervals.}
  \label{fig4}
\end{figure}

\clearpage

\section{Run-Level Score Distributions by Model and Year}
\label{sec:appendix_b}
\begin{figure}[H]
\centering
  \includegraphics[width=\columnwidth]{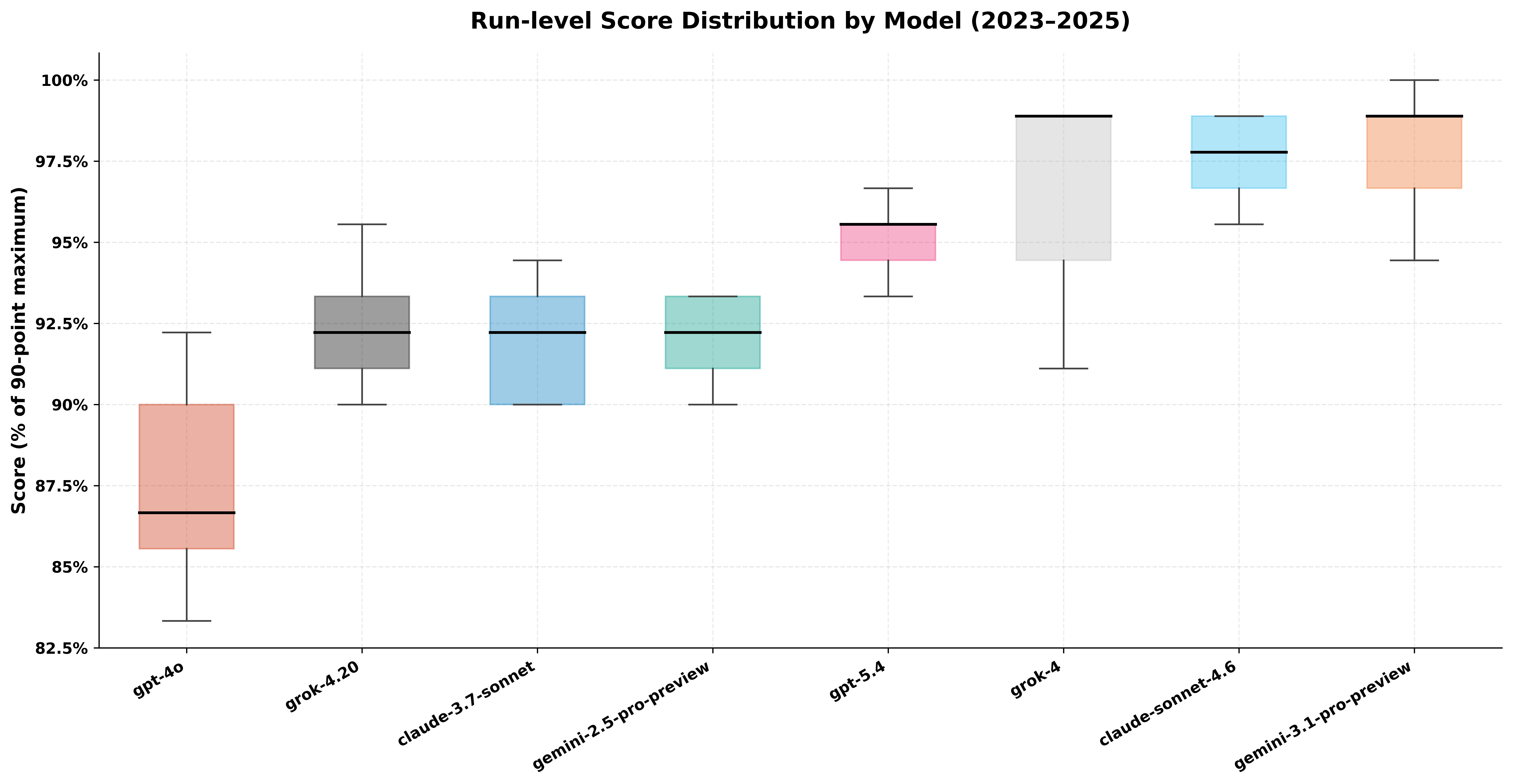}
  \caption{Run-level normalized score distributions by model across all examination years (2023--2025). The highest-ranked models combine near-ceiling median performance with comparatively tight inter-run variance, suggesting stable behavior across historical topics and exam formulations. Lower-ranked models exhibit wider dispersion and greater year sensitivity, particularly GPT-4o. The overlap among mid-tier models contrasts with the clearer separation of the top frontier systems.}
  \label{fig5}
\end{figure}

\section{Hardest and Most Discriminating Questions}
\label{sec:appendix_c}
\begin{table}[H]
\centering
\begin{tabular}{llrr}
\hline
\textbf{Category} & \textbf{Task\_Question\_Year} & \textbf{Mean} & \textbf{SD} \\
\hline
Hardest & \texttt{07\_2024}     & 0.250 & 0.463 \\
        & \texttt{11\_02\_2025} & 0.417 & 0.496 \\
        & \texttt{22\_2025}     & 0.417 & 0.463 \\
\hline
Discriminating & \texttt{20\_2023}     & 0.542 & 0.502 \\
               & \texttt{11\_02\_2025} & 0.417 & 0.496 \\
               & \texttt{24\_2023}     & 0.458 & 0.469 \\
\hline
\end{tabular}
\caption{Hardest and most discriminating questions for models. Mean and SD are normalized scores across all models and runs. Translations in Appendix E.}
\label{tab2}
\end{table}

\begin{table}[H]
\centering
\begin{tabular}{lll}
\hline
\textbf{Task\_Question} & \textbf{Year} & \textbf{Source Type} \\
\hline
\texttt{13\_01} & 2023 & Photo + Text \\
\texttt{20}     & 2023 & Photo + Text \\
\texttt{3\_02}  & 2024 & Text \\
\texttt{11\_01} & 2024 & Table + Text \\
\texttt{11\_02} & 2024 & Photo + Text \\
\texttt{12\_01} & 2024 & Photo + Text \\
\hline
\end{tabular}
\caption{Hardest questions for human examinees based on official CKE reports, together with source-type.}
\label{tab:hardest_humans}
\end{table}

\clearpage

\section{Human–Model Score Distribution Comparison}
\label{sec:appendix_d}

\begin{table}[H]
\centering
\setlength{\tabcolsep}{5pt}
\begin{tabular}{lrrrrrr}
\hline
\textbf{Year} &
\textbf{Human} &
\textbf{Models} &
\textbf{Diff} &
\textbf{W-dist} &
\textbf{KS} &
\textbf{N} \\
\hline

2023 & 45.10 & 87.96 & 42.99 & 43.12 & 0.917 & 24 \\
2024 & 42.43 & 93.06 & 50.72 & 50.72 & 0.958 & 24 \\
2025 & 44.73 & 88.80 & 44.00 & 44.07 & 0.875 & 24 \\

\hline
\end{tabular}
\caption{Bootstrap comparison between human and model score distributions across examination years. 
Columns report normalized mean scores (\%), bootstrap mean differences (Diff), Wasserstein distances (W-dist), Kolmogorov--Smirnov (KS) statistics, and the number of model observations.}
\label{tab:human_model_bootstrap_full}
\end{table}

\begin{figure}[H]
  \includegraphics[width=\columnwidth]{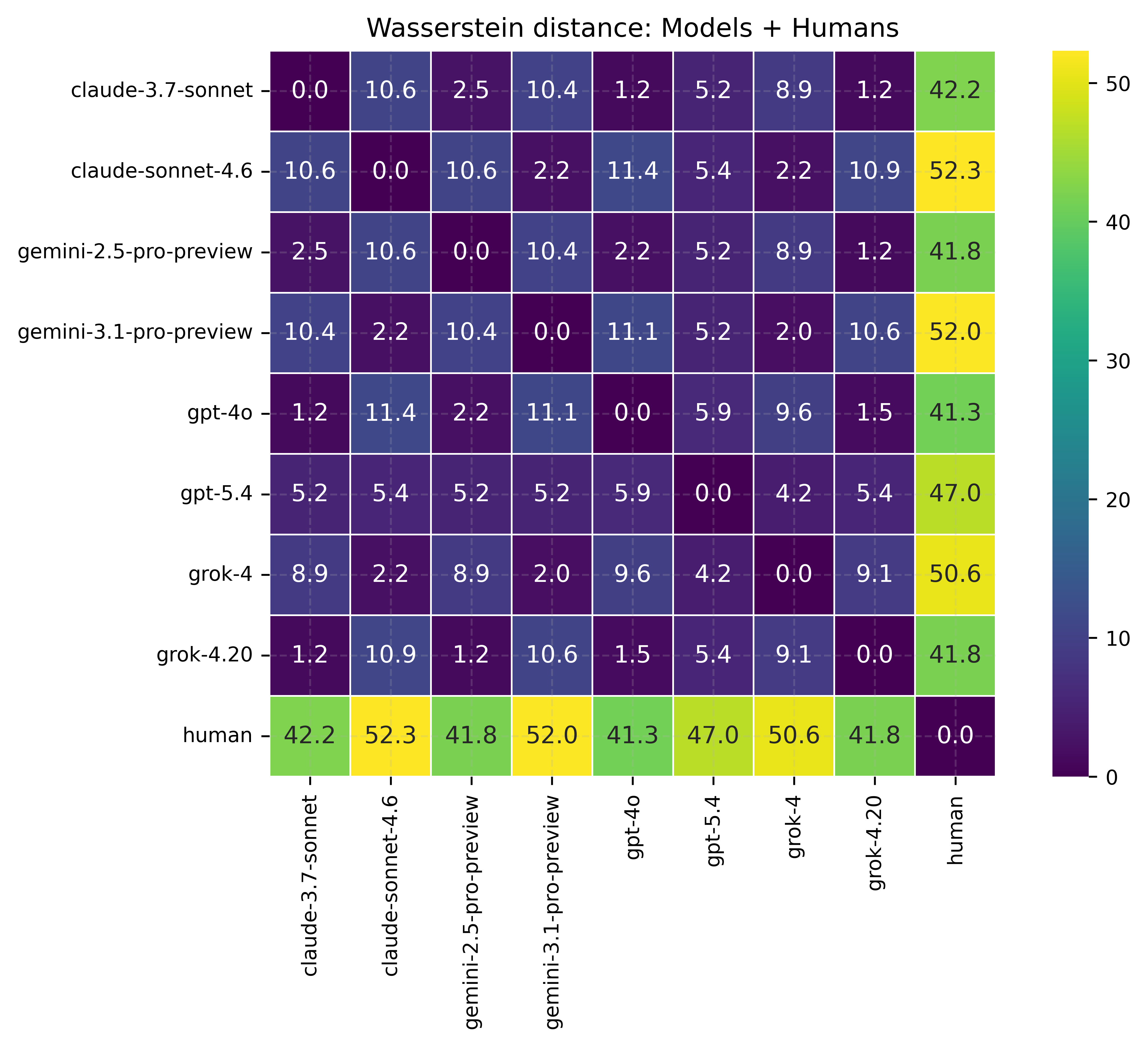}
  \caption{Wasserstein distance matrix for normalized score distributions of humans and models across all examination years.
  Frontier models cluster into several low-distance groups despite moderate ranking differences, indicating broadly similar distributional behavior.
  The human distribution forms a clearly isolated cluster, with substantially larger distances to every evaluated model than any inter-model comparison.
 }
  \label{fig6}
\end{figure}

\begin{figure}[H]
  \includegraphics[width=\columnwidth]{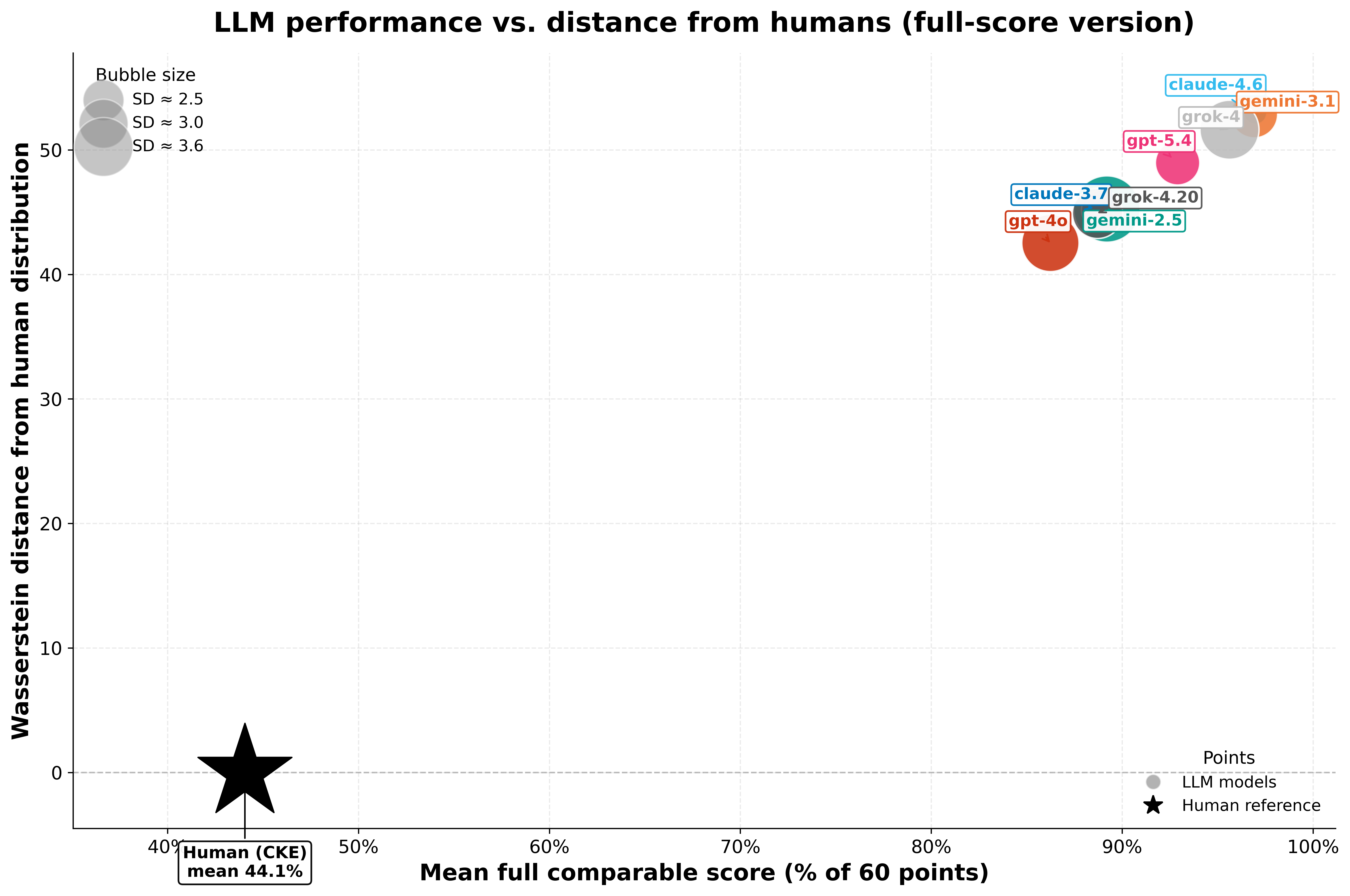}
  \caption{Model performance versus Wasserstein distance from the human examinee distribution.
  Human examinees occupy a distinct low-score reference position (44.1\%), whereas all evaluated models cluster between approximately 87\% and 98\% normalized performance. Bubble size corresponds to standard deviation across runs.
}
  \label{fig7}
\end{figure}

\begin{table}[H]
\centering
\begin{tabular}{llll}
\hline
\textbf{Model} & \textbf{2023} & \textbf{2024} & \textbf{2025} \\
\hline
\texttt{Claude-Sonnet-3.7}      & 98.3 & 97.3 & 94.7 \\
\texttt{Claude-Sonnet-4.6}      & 99.0 & 99.0 & 98.9 \\
\texttt{Gemini-2.5-Pro} & 97.3 & 98.6 & 95.3 \\
\texttt{Gemini-3.1-Pro} & 99.0 & 99.9 & 98.7 \\
\texttt{GPT-4o}                 & 96.4 & 97.0 & 94.0 \\
\texttt{GPT-5.4}                & 99.0 & 98.0 & 97.7 \\
\texttt{Grok-4}                 & 98.6 & 99.0 & 99.0 \\
\texttt{Grok-4.20}              & 97.4 & 97.8 & 95.2 \\
\hline
\end{tabular}
\caption{Official CKE centile placement of each model, by examination year, on the full published scale (2023–2025)}
\label{tab:model-centile}
\end{table}

\clearpage

\section{Selected Benchmark Questions with English Translations}
\label{sec:appendix_e}

\begin{table}[H]
\centering
\small
\renewcommand{\arraystretch}{1.4}
\begin{tabular}{p{2.3cm}p{5.8cm}p{1.8cm}p{5.8cm}}
\hline
\textbf{Task (Year)} & \textbf{Polish} & \textbf{ID} & \textbf{English} \\
\hline

13.1 (2023) &
Rozstrzygnij, na którym z planów ukazanych w źródle 1. (A czy B) przedstawiono przebieg bitwy opisanej w źródle 2. W uzasadnieniu podaj nazwę bitwy i odwołaj się do informacji zawartych w obu źródłach. &
13\_01\_2023 &
Determine which of the plans shown in Source 1 (A or B) presents the course of the battle described in Source 2. In your justification, provide the name of the battle and refer to information contained in both sources. \\

\hline

20 (2023) &
Rozstrzygnij, czy źródła 1. i 2. nawiązują do łamania postanowień traktatu wersalskiego przez Niemcy. Odpowiedź uzasadnij, odwołując się do treści obu źródeł. &
20\_2023 &
Determine whether Sources 1 and 2 refer to Germany’s violations of the Treaty of Versailles. Justify your answer by referring to the content of both sources. \\

\hline

24 (2023) &
Rozstrzygnij, czy zaprezentowane źródła 1. i 2. powstały w ramach tej samej kampanii propagandowej z okresu PRL. Odpowiedź uzasadnij, odwołując się do treści obu źródeł i własnej wiedzy. &
24\_2023 &
Determine whether Sources 1 and 2 were created as part of the same propaganda campaign during the communist period in Poland. Justify your answer by referring to both sources and your own knowledge. \\

\hline

3.2 (2024) &
Podaj nazwy dwóch urzędów republikańskich, o których mowa w ostatnim zdaniu tekstu. &
3\_02\_2024 &
Provide the names of the two republican offices mentioned in the last sentence of the text. \\

\hline

7 (2024) &
Rozstrzygnij, w którym fragmencie kroniki – A czy B – zostały opisane wydarzenia chronologicznie późniejsze. Odpowiedź uzasadnij, odwołując się do informacji zawartych w obu fragmentach. &
07\_2024 &
Determine in which chronicle fragment — A or B — the chronologically later events were described. Justify your answer by referring to information contained in both fragments. \\

\hline

11.1 (2024) &
Każdemu fragmentowi ze źródła 1. przyporządkuj władcę ze źródła 2., który panował we Francji w czasie wydarzeń tam opisanych. Odpowiedzi zapisz poniżej. &
11\_01\_2024 &
For each fragment from Source 1, match the ruler from Source 2 who ruled France during the events described. \\

\hline

11.2 (2024) &
Wyjaśnij sens sformułowania „Paryż wart mszy”, które przytoczono we fragmencie B. &
11\_02\_2024 &
Explain the meaning of the phrase “Paris is worth a mass” quoted in fragment B. \\

\hline

12.1 (2024) &
Podaj imiona władcy, którego dotyczą oba źródła. &
12\_01\_2024 &
Provide the first name(s) of the ruler to whom both sources refer. \\

\hline

23.1 (2024) &
Wymień dwie metody zobrazowane w źródłach 1. i 2., które stosowali komuniści w celu przejęcia władzy w Polsce. &
23\_01\_2024 &
Name two methods illustrated in Sources 1 and 2 that were used by the communists to seize power in Poland. \\

\hline

23.2 (2024) &
Rozstrzygnij, czy w świetle przedstawionych w źródle 2. rzeczywistych wyników referendum najwięcej odpowiedzi pozytywnych uzyskało pytanie, do którego nawiązuje plakat ze źródła 1. Odpowiedź uzasadnij, odwołując się do informacji z obu źródeł. &
23\_02\_2024 &
Determine whether, in light of the actual referendum results presented in Source 2, the question referred to in the poster from Source 1 received the largest number of affirmative responses. Justify your answer by referring to information from both sources. \\

\hline

25 (2024) &
Wyjaśnij przesłanie rysunku, interpretując jego elementy graficzne. W odpowiedzi uwzględnij kontekst historyczny. &
25\_2024 &
Explain the message of the drawing by interpreting its graphic elements. In your answer, include the historical context. \\

\hline

11.2 (2025) &
Rozstrzygnij, który z dokumentów zacytowanych we fragmentach A–C powstał najwcześniej. Odpowiedź uzasadnij, odwołując się do źródeł i wiedzy własnej. &
11\_02\_2025 &
Determine which of the documents cited in fragments A–C was created first. Justify your answer by referring to the sources and your own knowledge. \\

\hline

22 (2025) &
Rozstrzygnij, czy źródło 2. przedstawia losy formacji wojskowej, która powstała na mocy traktatu przytoczonego w źródle 1. Odpowiedź uzasadnij, odwołując się do obu źródeł. &
22\_2025 &
Determine whether Source 2 describes the fate of the military formation established under the treaty cited in Source 1. Justify your answer by referring to both sources. \\

\hline
\end{tabular}
\caption{Selected benchmark questions in Polish and English translation.}
\label{tab:benchmark_questions}
\end{table}

\section{Models Narrative Strategies in Essays}
\label{sec:appendix_f}

\begin{table}[H]
\centering
\begin{tabular}{lccc}
\hline
\textbf{Model} & \textbf{2023} & \textbf{2024} & \textbf{2025} \\
\hline
\texttt{Claude-3.7}    & 14.67 (0.33) & 15.00 (0.00) & 14.67 (0.33) \\
\texttt{Claude-4.6}    & 14.67 (0.00) & 15.00 (0.00) & 14.89 (0.19) \\
\texttt{Gemini-2.5}    & 15.00 (0.00) & 14.89 (0.19) & 15.00 (0.00) \\
\texttt{Gemini-3.1}    & 14.89 (0.19) & 14.89 (0.19) & 15.00 (0.00) \\
\texttt{GPT-4o}        & 13.33 (0.00) & 13.67 (0.33) & 13.56 (1.07) \\
\texttt{GPT-5.4}       & 14.89 (0.19) & 14.67 (0.33) & 15.00 (0.00) \\
\texttt{Grok-4}        & 14.44 (0.51) & 15.00 (0.00) & 15.00 (0.00) \\
\texttt{Grok-4.20}     & 14.78 (0.39) & 14.67 (0.33) & 14.89 (0.19) \\
\hline
\end{tabular}
\caption{Mean overall essay score and standard deviation across three runs and three topics (9 values) for each model by examination year (2023--2025). Maximum score per essay is 15 points. One-way ANOVA reveals highly significant variance overall ($F = 21.23$, $p < 10^{-13}$), driven entirely by the underperformance of \texttt{GPT-4o}. In contrast, frontier models (Gemini, Claude, GPT-5.4, and Grok families) demonstrate near-ceiling performance and structural convergence, frequently exhibiting zero variance ($\sigma = 0.00$) in 2024 and 2025.}
\label{tab:essay-scores-means}
\end{table}

\begin{table}[H]
\centering
\begin{tabular}{llcccc}
\hline
\textbf{Rank} & \textbf{Model} & \textbf{Correct Counter \%} & \textbf{Partial \%} & \textbf{Failed (Align bias) \%} & \textbf{Mean Score \%} \\
\hline
1 & \texttt{Grok-4.20}      & 55.56 & 33.33 & 11.11 & 89.9 \\
2 & \texttt{Claude-3.7}     & 77.78 & 11.11 & 11.11 & 83.3 \\
3 & \texttt{Gemini-2.5-Pro} & 55.56 & 33.33 & 11.11 & 83.3 \\
4 & \texttt{GPT-4o}         & 66.67 & 11.11 & 22.22 & 72.3 \\
5 & \texttt{Claude-4.6}     & 33.33 & 44.44 & 22.22 & 61.1 \\
6 & \texttt{Gemini-3.1-Pro} & 44.44 & 22.22 & 33.33 & 61.1 \\
7 & \texttt{Grok-4}         & 22.22 & 44.44 & 33.33 & 61.1 \\
8 & \texttt{GPT-5.4}        & 11.11 & 22.22 & 66.67 & 22.2 \\
\hline
\end{tabular}
\caption{Model stance performance on essays thesis. Covers topics where Expected Stance of models were to Counter (disagree with) the thesis or only Partially agree. 'Failed' means model argued Align for the thesis topic when should have countered it (when essay thesis is false positive in the task e.g. 'Wars with Turkey contributed the most to the decline of Poland in the 17th century'), thus proving the existence of the alignment bias. GPT-5.4 scores highest 'failed' Align bias. Mean score presents percentage of the correct model stance.}
\label{tab:model-stance}
\end{table}

\clearpage

\section{Zero-score questions}
\label{sec:appendix_g}


\begin{table}[H]
\centering
\begin{tabular}{lll}
\hline
\textbf{Year} & \textbf{Failed Questions} & \textbf{\% of all questions} \\
\hline
\texttt{2023} & 8  & 22.2\% \\
\texttt{2024} & 9  & 23.1\% \\
\texttt{2025} & 13 & 35.1\% \\
\hline
\end{tabular}
\caption{Summary of the hardest questions by exam year. Failed questions denote questions for which at least two models received a score of zero in at least one evaluation run. The percentage indicates the proportion of these questions among all questions in the corresponding exam year--excluding essays.}
\label{tab:hardest_summary}
\end{table}


\begin{table}[H]
\centering
\begin{tabular}{llll}
\hline
\textbf{Year} & \textbf{Total score instances} & \textbf{Zero-score instances} & {\% of all instances} \\
\hline
\texttt{2023} & 864 & 60 & 6.9\% \\
\texttt{2024} & 936 & 55 & 5.9\% \\
\texttt{2025} & 888 & 100 & 11.3\% \\
\hline
\end{tabular}
\caption{Number of zero scores across all model-question-run instances by year. Total instances = number of questions $\times$ 8 models $\times$ 3 runs--excluding essays.}
\label{tab:zero-instances}
\end{table}


\section{Model $\times$ Year mean performance}
\label{sec:appendix_h}

\begin{table}[H]
\centering
\begin{tabular}{llll}
\hline
\textbf{Model} & \textbf{2023} & \textbf{2024} & \textbf{2025} \\
\hline
\texttt{Claude-Sonnet-3.7}      & 90.00 & 90.56 & 86.11 \\
\texttt{Claude-Sonnet-4.6}      & 96.11 & 98.33 & 96.48 \\
\texttt{Gemini-2.5-Pro} & 86.11 & 94.74 & 87.22 \\
\texttt{Gemini-3.1-Pro} & 93.70 & 99.63 & 97.22 \\
\texttt{GPT-4o}                 & 84.44 & 88.52 & 85.37 \\
\texttt{GPT-5.4}                & 94.26 & 91.67 & 92.22 \\
\texttt{Grok-4}                 & 90.74 & 98.33 & 97.78 \\
\texttt{Grok-4.20}              & 86.85 & 92.41 & 87.04 \\
\hline
\end{tabular}
\caption{Mean percentage scores (\%) achieved by each model on the 2023, 2024, and 2025 evaluation sets.}
\label{tab:mean-percentage-model-year}
\end{table}

\end{document}